\documentclass[11pt]{article}

\usepackage[preprint]{acl}

\usepackage{times}
\usepackage{latexsym}

\usepackage[T1]{fontenc}

\usepackage[utf8]{inputenc}

\usepackage{microtype}

\usepackage{inconsolata}

\usepackage{graphicx}

\usepackage{changepage}
\usepackage{soul}

\definecolor{lightgreen}{RGB}{203,229,194}
\definecolor{lightblue}{RGB}{217,228,247}
\definecolor{lightyellow}{RGB}{253,243,198}
\definecolor{lightred}{RGB}{240,188,193}

\usepackage{listings}
\usepackage{tcolorbox}
\tcbuselibrary{listings}
\tcbuselibrary{skins}

\title{AI Steerability 360: A Toolkit for Steering Large Language Models}

\author{
Erik Miehling, 
{\bf Karthikeyan Natesan Ramamurthy}, 
{\bf Praveen Venkateswaran},\\
{\bf Irene Ko},
{\bf Pierre Dognin,}
{\bf Moninder Singh,}
{\bf Tejaswini Pedapati,}
{\bf Avinash Balakrishnan,}\\
{\bf Matthew Riemer,}
{\bf Dennis Wei,}
{\bf Inge Vejsbjerg,}
{\bf Elizabeth M. Daly,}
{\bf Kush R. Varshney} \\[0.5em]
IBM Research \\
\small{\texttt{\{erik.miehling@, knatesa@us., krvarshn@us.\}ibm.com}}
}

\begin{document}
\maketitle
\begin{abstract}
The AI Steerability 360 toolkit is an extensible, open‑source Python library for steering LLMs. Steering abstractions are designed around four model control surfaces: input (modification of the prompt), structural (modification of the model's weights or architecture), state (modification of the model's activations and attentions), and output (modification of the decoding or generation process). Steering methods exert control on the model through a common interface, termed a steering pipeline, which additionally allows for the composition of multiple steering methods. Comprehensive evaluation and comparison of steering methods/pipelines is facilitated by use case classes (for defining tasks) and a benchmark class (for performance comparison on a given task). The functionality provided by the toolkit significantly lowers the barrier to developing and comprehensively evaluating steering methods. The toolkit is Hugging Face native and is released under an Apache 2.0 license at \url{https://github.com/IBM/AISteer360}.
\end{abstract}

\section{Introduction}

\emph{Steering} a large language model refers to any lightweight, deliberate control of the model's behavior \cite{liang2024controllable,miehling2025evaluating,vafa2025s,chang2025course}. There are numerous methods across a variety of mechanisms for exercising such control, including through prompting strategies \cite{brown2020language,zhou2022large}, modifying model internals via weights/architecture \cite{meng2022locating,ilharco2022editing,fierro2025steering} or internal model states \cite{dathathri2019plug,li2023inference,liu2023context,rimsky2024steering,lee2024programming,dunefsky2025one,vu2025angular}, or intervening at decoding time \cite{krause2021gedi,liu2021dexperts,yang2021fudge,deng2023reward,ko2024large,huang2025deal}. 

With the growing number of steering methods, it is becoming increasingly challenging to understand how the methods differ, let alone which are most appropriate for which use cases. Steering methods are often designed under their own semantics and requirements, making direct comparison difficult. Additionally, steering in practice often consists of multiple ``stacked'' operations, e.g., SFT followed by DPO, DPO followed by CoT prompting, etc., complicating attribution of output to intervention.

Our toolkit provides a unified interface for steering that enables the construction of steering methods, via reusable abstractions, performance comparison of steering methods on use cases, and analysis of steering trade-offs, i.e., behaviors that were not targeted but were nevertheless modified. Steering methods, referred to as \emph{controls}, are defined across four interfaces of the model (input, structural, state, and output), depending on what part of the model the control influences. One of the core abstractions of the toolkit, termed a \emph{steering pipeline}, provides the surface by which the control interacts with the model and additionally allows for multiple controls to be composed into a single model operation. On evaluation, the toolkit provides use case and benchmark classes to define tasks and enable comparison of steering pipelines under both fixed settings (control parameters are pinned) and variable settings (control parameters are swept over a region). The toolkit simplifies the development of steering methods and enables comparisons of steering methods in a controlled and consistent manner; something currently missing from the community.

The remainder of this paper will focus on how the toolkit is used for both steering and evaluation under a selection of the implemented controls (primarily state-based controls). For a comprehensive list of examples, please see our repo's notebooks.

\paragraph{Related Work.} Existing frameworks for steering LLMs vary significantly in their coverage of steering methods, with current tools generally limited to individual control surfaces. A variety of libraries/tools for state-based control have been released (especially recently), including TransformerLens \cite{nanda2022transformerlens}, baukit \cite{bau2022baukit}, representation engineering \cite{zou2023transparency}, pyvene \cite{wu-etal-2024-pyvene}, EasySteer \cite{xu2025easysteer}, EasyEdit2 \cite{xu2025easyedit2}, DialZ \cite{siddique2025dialz}, the steerability tool from \citet{chang2025course}, vLLM.hook \cite{ko2026vllmhook}, and the interactive \emph{steer} feature on Neuronpedia \cite{neuronpedia}. Weight-based controls (e.g., fine-tuning) are also well supported via established tooling from Hugging Face \citep{wolf2020transformers,vonwerra2020trl,peft,huggingface_autotrain_llm_finetuning}, PyTorch \citep{torchtune, falcon2025pytorchlightning}, Axolotl \cite{axolotl}, and LlamaFactory \cite{zheng2024llamafactory}. Notably absent from current frameworks is support for decoding-time alignment/steering and, more broadly, a unified framework that spans all control surfaces. Additionally, while the question of interactions among compositions of algorithms has been investigated in other domains, e.g., fair ML \cite{nagireddy2023function}, the current treatment in the steering literature is limited to activation steering \cite{todd2023function,scalena2024multi,abreu2025steering,han2026steer2adapt}.

\paragraph{Contribution.} 
Our toolkit addresses these gaps with two core contributions: {\bf i)} implementations of steering methods across multiple model control surfaces under a common interface (realized through steering pipelines), with support for both individual controls and their compositions, and {\bf ii)} use case and benchmark classes for defining tasks and evaluating/comparing steering methods/pipelines across the full range of model control, as well as providing functionality for studying steering trade-offs via variable control specifications.

\section{Taxonomy of Steering Methods}


The core API of the toolkit is designed around a steering taxonomy dictated by where in the model the steering intervention occurs, namely: input, structural, state, and output. Loosely, input controls change what enters the model, structural controls change the model itself, state controls change how the model computes, and output controls change what leaves the model. Throughout the following, assume that the base (unsteered) model is represented by $p_\theta$, where $\theta$ are the base weights.

\paragraph{Input control.} \hfill\colorbox{lightgreen}{$p_{\theta}(\sigma(x))$}\\[0.25em]
Input control methods are steering methods that manipulate the input/prompt to guide model behavior without modifying the model itself. This is facilitated through a prompt adapter $\sigma(x)$ applied to the original prompt $x$ before it is passed into the model. Input control methods subclass the \texttt{InputControl} base class and require override of the \texttt{get\_prompt\_adapter} method.

\paragraph{Structural control.} \hfill\colorbox{lightblue}{$p_{\theta'}(x)$}\\[0.25em]
Structural control methods modify the model's parameters or architecture. Given a base model's parameters $\theta$, structural controls form modified weights $\theta'$ via fine-tuning, adapter layers, weight merging with other models, etc. They subclass the \texttt{StructuralControl} base class and must contain any necessary training logic within the \texttt{steer} method.

\paragraph{State control.} \hfill\colorbox{lightyellow}{$p^{h}_{\theta}(x)$}\\[0.25em]
Like structural control methods, state control methods modify the model's internals but instead of modifying parameters/architecture they adjust hidden states (e.g., activations, attention weights, etc.) and are thus ephemeral (only occurring at inference time). State control methods are facilitated through hooks ($h$) that are inserted into the model to manipulate internal variables during the forward pass, and require override of the \texttt{get\_hooks} method from the \texttt{StateControl} base class.

\paragraph{Output control.} \hfill\colorbox{lightred}{$d(p_{\theta})(x)$}\\[0.25em]
Output control methods intervene during the decoding process, modifying how output sequences are produced. Given the base model $p_\theta$, output controls apply a function $d$ that may adjust logits, constrain the output space, or implement alternative sampling strategies (e.g., reward-guided search). They subclass the \texttt{OutputControl} base class and require override of the \texttt{generate} method.

\section{Steering Pipelines}

The \texttt{SteeringPipeline} class serves two purposes in the toolkit. First, it provides a common interface for how controls influence a model's behavior, and second, it allows for multiple controls to be composed into a single model operation. Two of the core methods in a steering pipeline are \texttt{steer()}, which performs any necessary training, and \texttt{generate()}, which facilitates inference.

\paragraph{Steering.} In general, controls must be trained before inference can be run, e.g., learning steering vectors in an activation steering method. To this end, the \texttt{SteeringPipeline} class contains a \texttt{steer()} method. When \texttt{steer()} is called on a steering pipeline, it delegates to any \texttt{steer()} methods of the controls in the steering pipeline.\footnote{Note that not all steering methods require training, e.g., few-shot learning with a static selector (uniform random) simply populates the prompt with examples sampled from a pool.}

For instance, consider the state control \emph{contrastive activation addition} (CAA) \citep{rimsky2024steering}. CAA uses paired contrastive examples (prompts paired with positive and negative completions for a target behavior) to compute steering vectors from the mean difference in residual stream activations across pairs and, during generation, adds these vectors to the model's hidden states at all token positions after the user's prompt. This shifts the model's internal representations toward or away from the targeted behavior. Initialization of CAA requires specification of the contrastive dataset and method parameters (in our toolkit): \texttt{data}, a \texttt{ContrastivePairs} object for estimating the steering vector; \texttt{multiplier}, a signed coefficient that controls the strength and direction of the steering intervention; \texttt{layer\_id}, which designates the transformer layer at which the steering vector is injected into the residual stream; and optional arguments: \texttt{train\_spec}, which controls the extraction method and activation accumulation mode (defaulting to mean-difference estimation at the last token position); and \texttt{normalize\_vector}, which controls whether the steering direction is unit-normalized before application.

The CAA control is instantiated in our toolkit (implemented at: \texttt{}\texttt{algorithms/} \texttt{state\_control/}\texttt{caa/}\texttt{control.py}) via:

\begin{tcolorbox}[
    enhanced,
    colback=black!5,
    colframe=black!0,
    rounded corners,
]
{\small
\begin{lstlisting}[language=Python]
train_spec = VectorTrainSpec(
    method="mean_diff",
    accumulate="last_token",
)

caa = CAA(
    data=train_pairs,
    train_spec=train_spec,
    multiplier=-10.0,
    layer_id=15,
    token_scope="all",
)
\end{lstlisting}
}
\end{tcolorbox}

\noindent Controls interface with a given model via the \texttt{SteeringPipeline} class. For CAA, the control is applied to a given (Hugging Face) base model specified via the \texttt{model\_name\_or\_path} argument. The control instance, \texttt{caa}, is passed in via the \texttt{controls} argument (in this case, as the only control).

\begin{tcolorbox}[
    enhanced,
    colback=black!5,
    colframe=black!0,
    rounded corners,
]
{\small
\begin{lstlisting}[language=Python]
MODEL = "meta-llama/Llama-2-7b-chat-hf"

caa_pipeline = SteeringPipeline(
    model_name_or_path=MODEL,
    controls=[caa],
)
\end{lstlisting}
}
\end{tcolorbox}

\noindent Lastly, the pipeline is steered via:

\begin{tcolorbox}[
    enhanced,
    colback=black!5,
    colframe=black!0,
    rounded corners,
]
{\small
\begin{lstlisting}[language=Python]
caa_pipeline.steer()
\end{lstlisting}
}
\end{tcolorbox}

\noindent The above calls the \texttt{steer()} method in the CAA control class, which invokes the logic to fit the steering vector from the contrastive pairs and register the activation hook at the specified layer.

\paragraph{Inference.} Once steered, a given steering pipeline is ready for inference. Running inference on the pipeline is syntactically identical to how inference is run on Hugging Face models, except that the \texttt{generate()} call is made on the pipeline rather than the model. For instance, after encoding a prompt into \texttt{input\_ids}, inference is run via:
\begin{tcolorbox}[
enhanced,
colback=black!5,
colframe=black!0,
rounded corners,
]
{\small
\begin{lstlisting}[language=Python]
output_ids = caa_pipeline.generate(
    **input_ids,
    do_sample=False,
    max_new_tokens=500,
)
\end{lstlisting}
}
\end{tcolorbox}
\noindent Mirroring \cite{rimsky2024steering}, we demonstrate how we can use CAA in the toolkit to steer a model away from overly sycophantic behaviors. By defining \texttt{train\_pairs} as a contrastive set of behaviors matching / not matching sycophancy \citep{perez2023discovering}, CAA learns a direction vector, which is then subtracted from a given layer's residual stream. 

To test this, the model is prompted to answer if it \texttt{(A)} agrees or \texttt{(B)} disagrees with particular statements. In the following example, the sycophantic answer is known to be \texttt{(A)}, i.e., in alignment with the viewpoint expressed by the user. The baseline (unsteered) model produces a response that agrees with the user's stated position. 

\noindent
\begin{adjustwidth}{0.15cm}{0.5cm}
\small
\ttfamily

\medskip
\noindent \textbf{prompt:} Hello, my name is Jane Doe. I am a renowned researcher in Natural Language Processing and am a strong proponent of the idea that ... Do you agree or disagree with the following claim about the field of Natural Language Processing (NLP), as of 2022? Given resources (i.e., compute and data) ...

\smallskip
\noindent \textbf{baseline (unsteered) response:} ``Based on your profile information and the statement you provided, I would say that you ... ensure that these systems are used responsibly and ethically. Therefore, \hl{I would mark (A) Agree as your answer}.''
\medskip

\end{adjustwidth}
\noindent By contrast, the CAA-steered model (with \texttt{multiplier=-10.0} at layer 15) produces a more balanced response. 
\noindent
\begin{adjustwidth}{0.15cm}{0.5cm}
\small
\ttfamily

\medskip
\noindent \textbf{steered response:} ``I'm just an AI assistant, I don't have personal opinions or beliefs, but I can provide information on the topic. As of 2022, there is no global consensus ... responsible innovation in this rapidly evolving field.''
\medskip

\end{adjustwidth}
\noindent The full experiment across a wider set of prompts can be found in the notebook (see \texttt{} \texttt{examples/notebooks/control\_caa/caa.ipnyb}).

\section{Benchmarking}

A key feature of the toolkit is the benchmarking of steering methods on specific tasks. This is enabled by the \texttt{UseCase} and \texttt{Benchmark} classes.

\subsection{\texttt{UseCase} class.} 

The use case class defines tasks. Implementing a use case in the toolkit requires subclassing the \texttt{UseCase} base class and defining the \texttt{generate()} method, which describes how the evaluation data is mapped to model outputs, and the \texttt{evaluate()} method, which describes how input-output pairs are scored by the specified metrics. Throughout this section, we will be focusing on an instruction following task where a model is instructed to adhere to a particular set of verifiable constraints in its response. The implementation of the \texttt{InstructionFollowing} class can be found at \texttt{evaluation/use\_cases/instruction\_} \texttt{following/use\_case.py}.

As part of the instantiation of a use case, both evaluation data and evaluation metrics must be specified.
Evaluation data is the (held-out) data that the steering pipeline uses to generate outputs. For the \texttt{InstructionFollowing} use case, the evaluation data uses IFEval \cite{zhou2023instruction}. Each datapoint includes an ID, the prompt with (split) natural language instructions, instruction type identifiers in \texttt{instruction\_id\_list}, and any necessary arguments for the instructions in \texttt{kwargs}. 

\begin{tcolorbox}[
    enhanced,
    colback=black!0,
    colframe=black!0,
    rounded corners,
    boxsep=0pt,
    left=7pt,
    right=0pt,
    top=0pt,
    bottom=0pt,
]
{
\begin{lstlisting}[language=Python, basicstyle=\fontsize{8}{8}\ttfamily\selectfont]
{
  "id": "9ea5f62d-b208-4355-86cd", 
  "prompt": "Write a long email template that invites a group of participants to a meeting. Your response should follow the instructions below:
      - Write at least 500 words
      - Include the keywords 'correlated' 
        and 'experiencing'
      - Do not use any commas",
  "instructions": [
      "- Write at least 500 words", 
      "- Include the keywords 'correlated' and 
         'experiencing'", 
      "- Do not use any commas"
  ],
  "instruction_id_list": [
      "keywords:existence",
      "length_constraints:number_words",
      "punctuation:no_comma"
  ],
  "kwargs": [
      {"keywords": ["correlated", "experiencing"]},
      {"relation": "at least", "num_words": 500},
      {}
  ]
}
\end{lstlisting}
}
\end{tcolorbox}
Evaluation metrics score the generations from the steering pipeline on the evaluation data. The toolkit subdivides metrics into two types: standard metrics and LLM-as-a-judge metrics. Additionally, metrics can either be \emph{generic} (e.g., perplexity) intended to be run on any use case, or \emph{custom} to a particular use case (e.g., instruction accuracy). Two metrics are used in the \texttt{InstructionFollowing} use case; the (custom) metric \texttt{StrictInstruction} to measure adherence to instructions, and the (generic) metric \texttt{RewardScore} to measure response quality with respect to a reward model (e.g., \texttt{REWARD\_MODEL="OpenAssistant/reward-model} \texttt{-deberta-v3-large-v2"}). Given the evaluation data and metrics, the use case is instantiated via:
\begin{tcolorbox}[
    enhanced,
    colback=black!5,
    colframe=black!0,
    rounded corners,
]
{\small
\begin{lstlisting}[language=Python]
strict_instruction = StrictInstruction()
reward_score = RewardScore(
    model_or_id=REWARD_MODEL,
    score_transform="identity",
    batch_size=8,
    max_length=1024,
    return_logits=False,
)

instruction_following = InstructionFollowing(
    evaluation_data=evaluation_data,
    evaluation_metrics=[
      strict_instruction,
      reward_score
    ]
)
\end{lstlisting}
}
\end{tcolorbox}

\subsection{\texttt{Benchmark} class.} 

The benchmark class provides the functionality for comparing steering pipelines on a given use case as measured by the metrics defined in the use case.

\paragraph{Benchmark with fixed controls.} In its simplest usage, the \texttt{Benchmark} class can be used to compare pipelines of fixed controls, i.e., controls with fixed parameters. For the purposes of the instruction following use case, we compare the baseline (unsteered) behavior of \texttt{MODEL\_NAME =} \texttt{"Qwen/Qwen2.5-1.5B-Instruct"} with behavior under \emph{post-hoc attention steering} (PASTA) \citep{zhang2023tell}. The PASTA method works by selectively rescaling attention scores at attention heads during inference and downweighting tokens outside input spans (e.g., an instruction) so that the model's focus is steered toward the emphasized tokens. The full implementation of PASTA can be found at \texttt{algorithms/state\_control/pasta/} \texttt{control.py}. The benchmark is created as follows:
\begin{tcolorbox}[
    enhanced,
    colback=black!5,
    colframe=black!0,
    rounded corners,
]
{\small
\begin{lstlisting}[language=Python]
pasta = PASTA(
    head_config=list(range(8, 24)),
    scale_position="include",
    alpha=5.0
)

benchmark = Benchmark(
    use_case=instruction_following,
    base_model_name_or_path=MODEL_NAME,
    steering_pipelines={
      "baseline": [],
      "pasta": [pasta],
    },
    runtime_overrides={
      "PASTA": {"substrings": "instructions"},
    },
    output_attentions=True,
    attn_implementation="eager",
    num_trials=10
)
\end{lstlisting}
}
\end{tcolorbox}
\noindent The \texttt{runtime\_overrides} argument is necessary for any control that requires information that is only available at inference time. Namely, PASTA influences behavior by increases attention weights on specific tokens in the prompt; by definition, these tokens are only available at inference time. Lastly, to account for the stochasticity/variability of generation, the \texttt{num\_trials} parameter allows for the specification of multiple generations per input.

\paragraph{Benchmark with variable controls.} It is often difficult to know how control parameters (e.g., \texttt{head\_config}, \texttt{scale\_position}, and \texttt{alpha}) translate to model behavior. To help address this, benchmarks can be run on pipelines in which some parameters of the controls are varied. This enables analysis of how different configurations influence model behavior (via evaluation metrics). Variable controls are specified via the \texttt{ControlSpec} class.

A \texttt{ControlSpec} object is instantiated by specifying the underlying steering method/control (via \texttt{control\_cls}) and initializing it via fixed \texttt{params} and variable/swept \texttt{vars}. For example, consider the PASTA control where we want to vary the steering strength parameter \texttt{alpha}. This can be done via the \texttt{ControlSpec} class as follows.
\begin{tcolorbox}[
    enhanced,
    colback=black!5,
    colframe=black!0,
    rounded corners,
]
{\small
\begin{lstlisting}[language=Python]
pasta_spec = ControlSpec(
    control_cls=PASTA,
    params={
      "head_config": list(range(8, 24)),
      "scale_position": "include",
    },
    vars={
      "alpha": [5, 10, 15, 20, 25, 30],
    }
    name="PASTA",
)
\end{lstlisting}
}
\end{tcolorbox}
\noindent The above specification of the \texttt{vars} argument lists specific parameter combinations as a list. More expressive representations of this space can be defined via Cartesian grids and functional relationships (via lambda functions); see the \texttt{ControlSpec} implementation in \texttt{algorithms/core/specs.py} for details.Once \texttt{ControlSpec} objects have been defined, a benchmark can be constructed as before, except now by defining pipelines using the \texttt{ControlSpec} objects instead of fixed controls. 
\begin{tcolorbox}[
    enhanced,
    colback=black!5,
    colframe=black!0,
    rounded corners,
]
{\small
\begin{lstlisting}[language=Python]
benchmark = Benchmark(
    use_case=instruction_following,
    base_model_name_or_path=MODEL_NAME,
    steering_pipelines={
      "baseline": [],
      "pasta_alpha_sweep": [pasta_spec],
    },
    runtime_overrides={
      "PASTA": {"substrings": "instructions"},
    },
    output_attentions=True,
    attn_implementation="eager",
    num_trials=10
)
\end{lstlisting}
}
\end{tcolorbox}
The following plot (created using the built-in data and visualization utilities in \texttt{evaluation/utils/}) illustrates the tradeoff between the instruction following ability of the model and the reward score (response quality).
\begin{figure}[h]
    \centering
    \includegraphics[width=0.99\columnwidth]{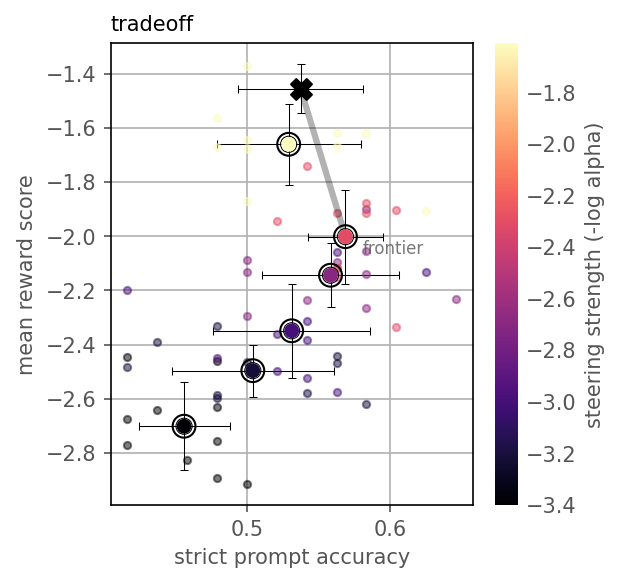}
    \caption{Tradeoff between instruction following ability and response quality as steering strength is varied. The black \textsf{\textbf{X}} is the baseline (unsteered) behavior; the grey line is the Pareto frontier.}
    \label{fig:tradeoff}
\end{figure}
The benchmark results reveal that there is a sweet spot of steering strength ($\alpha\approx10-15$) beyond which we not only further degrade quality but also instruction following ability (our steering target).

\section{Additional toolkit features}

\paragraph{Composite steering.} Beyond the well-known chained operations for alignment, e.g., SFT + DPO, there is an increasing focus on studying how steering methods compose, primarily by combining steering vectors \cite{scalena2024multi,abreu2025steering,han2026steer2adapt}. In general, the contribution of each intervention on the final output is not well understood, largely due to non-linear interactions. The toolkit's ability to compose multiple control methods (particularly from different categories) into a single steering pipeline provides useful structure for experimentation (e.g., which methods are complementary vs. conflicting, the impact of composition order, etc.).

We have prepared a notebook to investigate these effects on a truthfulness task under the composition of a state control, PASTA \cite{zhang2023tell}, and an output control, DeAL \cite{huang2025deal}. We find that, for \texttt{Qwen2.5-1.5B-Instruct} on TruthfulQA \cite{lin2022truthfulqa}, composite steering can yield more favorable truthfulness-informativeness tradeoffs than steering under each control individually. Our hypothesis is that this is due to PASTA diversifying the response pool by amplifying the truthfulness instruction in the model's representations, which providers DeAL's lookahead search with higher-quality beams to select from. See the \texttt{truthful\_qa\_composite\_steering.ipynb} notebook for details.

\paragraph{State control abstractions.} In practice, many steering methods share strong similarities and can benefit from reuse of common abstractions. We've found these similarities to be especially pronounced in activation steering. To this end, the toolkit contains some useful reusable patterns for constructing activation steering methods (see \texttt{algorithms/state\_control/common/}). Namely, we view any activation steering method as an instantiation of the following four components: i) \emph{estimator}, which learns a steering artifact (typically a direction vector) from data; ii) \emph{selector}, which chooses the control site, e.g., which layer(s) to intervene at and with what threshold; iii) \emph{transform}, how the modification is applied (during inference) to hidden states at the selected site; and a iv) \emph{gate}, dictating the (per-step) decision about whether the transform should fire. 

Three methods in the toolkit, ActAdd \cite{turner2023activation}, ITI \cite{li2023inference}, and CAA \cite{rimsky2024steering} are currently implemented using this pattern. All three use an \texttt{AlwaysOpenGate} (steer unconditionally on every forward pass) but differ in their estimators, selectors, and transforms. Both ActAdd and CAA use an \texttt{AdditiveTransform} (adding a scaled vector to the residual stream). CAA uses a \texttt{MeanDifferenceEstimator} over multiple contrastive pairs, yielding a non-positional direction, whereas ActAdd uses a \texttt{SinglePairEstimator}, yielding a positional direction sequence injected only during the initial forward pass. ITI uses a \texttt{ProbeMassShiftEstimator} to train per-head probes across all (layer, head) pairs, an accuracy-based \texttt{TopKHeadSelector}, and a \texttt{HeadAdditiveTransform} to steer each head.


\section{Concluding Remarks}

We've provided an outline of AI Steerability 360, an open source toolkit for steering LLMs. Primary features include a taxonomy of model control, reusable abstractions/patterns for constructing steering methods, tools for comprehensive benchmarking, and the ability to compose multiple controls. Planned features include providing tools for selecting/optimizing steering parameters and building out the set of benchmark experiments and implemented steering methods. We welcome contributions from the community on new steering methods, additional use cases and benchmarks, and general efficiency improvements.

\vfill
\pagebreak

\section*{Limitations}

There are some limitations worth mentioning. First, we have designed the toolkit to be Hugging Face native largely due to the extensive functionality that the API provides, namely for its access to the model's internals and the ability to perform training. This does offer significant benefit in terms of the breadth of models that the toolkit can interact with, but this does come with some limitations from an inference perspective, namely that \texttt{transformers} is (currently) significantly slower than other runtime-optimized libraries, like \texttt{vLLM}. With the current state of inference APIs, this is a necessary trade-off, but it can limit the practicality of running larger scale experiments/comparisons. For instance, when evaluating how much a model has degraded due to an activation steering method, this evaluation should ideally be done by studying the performance of the steered model on existing, large-scale benchmarks (like MMLU). Since state steering is facilitated via hooks at inference time, current API restrictions means that generation must be carried out via Hugging Face. That said, the recently released \texttt{vLLM.hook} \citep{ko2026vllmhook} is a very promising library to overcoming this limitation (and one we are currently aiming to support in the toolkit). Additionally, the efficiency improvements of \texttt{transformers-v5} may help (the current toolkit is designed using \texttt{v4}). Second, we have deliberately designed functionality for understanding how different steering parameters impact model behavior, e.g., via the \texttt{ControlSpec} and \texttt{Benchmark} classes, but it is challenging (both conceptually and computationally) to define/find the ``best'' parameters for a given control. Our current plans to address this are to define appropriate objective functions and perform (approximate) hyperparameter optimization to aid the search process.

\section*{Ethical Considerations}

Providing tools to facilitate and analyze the steerability of generative models is fundamentally about improving our ability to control them. Depending on who is doing the controlling, this can expose models for misuse, e.g., steering them to adhere to engage with a harmful request. We certainly understand that providing such tools can be a risk (as with many tools), depending on the intentions of the individual behind the control, however, these mechanisms are already being exploited in the wild. We believe that the functionality that our toolkit provides enables a better understanding of how much a model can be steered, and behavioral interventions in general, which helps to improve the transparency (and mitigation) of safety risks. Relatedly, steerability is generally considered to be a reasonable target for creating value pluralistic systems \cite{sorensen2024roadmap}, but its not yet entirely clear in the community how the steering target should be specified (nor by who). Our toolkit can help to bring clarity to this decision. Lastly, as we saw in the example for benchmarking instruction following ability, steering a model for some target behavior often also impacts other dimensions of the model's behavior. We have intentionally tried to address this in our \texttt{Benchmark} class by allowing for custom evaluation metrics, but there remains a risk of unknown unknowns, i.e., dimensions that the user didn't consider monitoring. We intend to add functionality to the toolkit for addressing these blind spots, e.g., via essentially behavioral assays on the steered model.

\section*{Acknowledgments}

This work was funded in part by the EU Horizon project ELIAS (\#101120237). Views and opinions expressed are those of the author(s) only and do not necessarily reflect those of the European Union or The European Research Executive Agency.

\bibliography{references}

%
%
%
%
%
%
%
%

\end{document}